# Improving Generalizability of Extracting Social Determinants of Health Using Large Language Models through Prompt-tuning

Cheng Peng, PhD[1], Zehao Yu, MS[1], Kaleb E Smith, PhD[3], Wei-Hsuan Lo-Ciganic[4], Jiang Bian, PhD[1,2], Yonghui Wu, PhD[1,2]

[1]Department of Health Outcomes and Biomedical Informatics, College of Medicine, University of Florida, Gainesville, Florida, USA; [2]Cancer Informatics Shared Resource, University of Florida Health Cancer Center, Gainesville, Florida, USA; [3]NVIDIA, Santa Clara, California, USA; [4]Department of Pharmaceutical Outcomes & Policy, College of Pharmacy, University of Florida, Gainesville, Florida, USA

**Abstract**

*The progress in natural language processing (NLP) using large language models (LLMs) has greatly improved patient information extraction from clinical narratives. However, most methods based on the fine-tuning strategy have limited transfer learning ability for cross-domain applications. This study proposed a novel approach that employs a soft prompt-based learning architecture, which introduces trainable prompts to guide LLMs toward desired outputs. We examined two types of LLM architectures, including encoder-only GatorTron and decoder-only GatorTronGPT, and evaluated their performance for the extraction of social determinants of health (SDoH) using a cross-institution dataset from the 2022 n2c2 challenge and a cross-disease dataset from the University of Florida (UF) Health. The results show that decoder-only LLMs with prompt tuning achieved better performance in cross-domain applications. GatorTronGPT achieved the best F1 scores for both datasets, outperforming traditional fine-tuned GatorTron by 8.9% and 21.8% in a cross-institution setting, and 5.5% and 14.5% in a cross-disease setting.*

**Introduction**

Natural language processing is the key technology to extract patient information from clinical narratives, which contain rich information about patients' history pivotal to medical understanding and healthcare delivery. However, many studies[1–5] have discovered large performance drops due to the documentation variations in cross-institution settings, which is a barrier to adopting clinical text analytics in multi-site studies. To account for the performance drop, researchers must annotate additional data and retrain algorithms for cross-institution studies, which blocks many institutions from adopting the power of clinical text analytics. Transfer learning,[6] the ability to learn knowledge from one task and apply it to another task (often involving data from a different resource), is a promising solution to improve the portability of NLP for cross-institution applications. Recent studies show that transformer-based large language models (LLMs) have good abilities in few-shot learning (learning from very few labeled samples),[7] zero-shot learning (learning without labeled samples for the target application),[8] and transfer learning[9] to enable better language processing. Transformer-based LLMs perform better by using self-supervised pretraining (training LLMs using unlabeled text) and supervised fine-tuning (i.e., applying LLMs to solve specific tasks).[10] After successful pretraining, one LLM can be used to solve many NLP subtasks through fine-tuning (i.e., transfer learning).[6] Recent studies show that LLMs based on the Generative Pre-trained Transformer (GPT) architecture[11] (e.g., GPT-3[7]) demonstrate significantly better transfer learning ability across different domains compared with traditional NLP models.

Most current NLP systems for patient information extraction are implemented using a two-stage solution including clinical concept extraction and relation extraction.[12–14] Clinical concept extraction is to identify specific medical terms or entities from the text (e.g., diseases, treatments, and symptoms); relation extraction is to establish semantic relationships among these extracted concepts, e.g., a drug causes a specific adverse drug event. Various solutions including rule-based, machine learning-based, and hybrid systems have been developed. Most state-of-the-art NLP systems are based on machine learning techniques (e.g., the conditional random fields[15] and support vector machines[16]), and more recently, by deep learning models, such as convolutional neural network[17], recurrent neural network[18] and its variants long short-term memory network[19], until the recent transformer-based LLMs. Inspired by the self-attention mechanism[20], transformer models have become the state-of-the-art methods for NLP, such as BERT[21], ALBERT[22], RoBERTa[23] and ELECTRA[24], given their ability to handle long-term dependencies and capture complex semantic information in clinical texts. Yang et al.[10,25] systematically explored four transformer architectures for clinical concept extraction and three transformer architectures for clinical relation extraction, demonstrating the effectiveness of transformers for clinical NLP tasks. Most studies exploring transformer models are based on fine-tuning, where all parameters of the transformer model are updated. Recent studies show that the generalizability of transformer models with fine-tuning is challenging for cross-institution applications, limiting the adoption of text analytics in cross-institution applications.[26] For instance, in the 2022 n2c2 challenge on social determinants of health (SDoH), the top-performing teams showed significant performance drops of 7~17% when applying transformer models fine-tuned using MIMIC data to a testing dataset developed using clinical notes from the University of Washington (UW).[27] Even if the training data and testing data are from the same institution, the portability of fine-tuned transformer models is still challenging when applied to a different disease domain, due to the variations in disease-specific terminologies, diagnostic criteria, and treatment protocols. Our previous study shows that LLMs fine-tuned using cancer patients' notes suffered a performance drop when applied to another cohort with opioid use. To address this challenge, several domain adaptation and transfer learning approaches have been proposed. For example, data augmentation or selection,[28] combining or transferring model parameters between domains.[29,30]

Most recently, prompt-based learning algorithms have shown promising transfer learning and few-shot learning capabilities, which could be applied to improve the portability of clinical NLP systems for patient information extraction.[31] Prompt-based learning framework reformulates the downstream task by adding a prompt to the input instead of adapting pre-trained language models to downstream tasks via the task-specific output layer.[32] There are two main types of prompts[32] including (1) "hard prompts", which are predefined textual inputs crafted by humans, the parameters of LLMs are updated during hard prompt-based learning; and (2) "soft prompts", which are trainable embeddings that added into the input to control the model's learning behavior. An efficient way to use soft prompts is to keep the LLM parameters unchanged and only update the parameters of the soft prompts, which is known as prompt-tuning (i.e., P-tuning). Both hard and soft prompt-based learning have been successfully applied to clinical NLP tasks. In a recent study, we successfully formulated the extraction of clinical concept and relation as a reading comprehension (MRC) task and explored two types (hard and soft) of prompts in multiple clinical transformers, demonstrating the efficiency of P-tuning for cross-institution application.[33] However, in our previous study, we only explored encoder-only LLMs, where task-specific layers are needed to adopt LLMs for various downstream tasks.

In this study, we seek to examine the transfer learning ability of LLMs using the P-tuning framework for extraction of SDoH from clinical notes. We examined two major types of LLM architectures, including the encoder-only architecture - designed to encode the input text into a high-dimensional space and understand its semantic representation, and the decoder-only architecture - designed to predict the next word in a sequence given the previous text. We compared our methods with previous non-prompt transformers using two clinical benchmark datasets focusing on SDoH, including the 2022 n2c2 challenge dataset for cross-institution adaptation and the UF Health notes for cross-disease adaptation. The experimental results show that P-tuning of generative LLMs improved clinical concept extraction and relation extraction of SDoH for cross-institution and cross-disease settings.

**Methods**

**Dataset**

This study used two datasets focusing on SDoH extraction from clinical notes. The first dataset is developed by the 2022 n2c2 challenge track 2,[27] where the Social History Annotated Corpus (SHAC) is used for model training and evaluation. SHAC includes 4,405 social history sections (notes) from the MIMIC-III database and clinical notes of UW, where annotators annotated 5 categories of SDoH concepts and 9 categories of SDoH-associated attribute concepts (e.g., Frequency, Type), and 28 categories of relations between SDoH concepts and SDoH-associated attributes. We examined the transfer learning ability of LLMs for cross-institution applications by fine-tuning LLMs using the MIMIC-III training set and evaluating performance using the UW test set. The second SDoH dataset is developed using clinical notes from the University of Florida (UF) Health Integrated Data Repository (IDR). We collected two disease cohorts including a cancer cohort and a opioid use cohort and identified 19 categories of SDoH concepts and 26 categories of relations between SDoH concepts from the clinical notes. Detailed information about this dataset is described in our previous study[34]. We examined the transfer learning ability for cross-disease domain applications by fine-tuning LLMs using the cancer SDoH dataset and evaluating performance using the opioid SDoH dataset. **Table 1** shows the distribution of notes and SDoH concepts and relations in the two datasets.

**Table 1.** Summary statistics of the clinical notes and annotated concepts and relations in the 2022 n2c2 challenge dataset and the UF Health notes dataset.

| Corpus | Datasets | Number of notes | Number of clinical concepts | Number of clinical relations |
|---|---|---|---|---|
| 2022 n2c2 challenge | MIMIC-training | 1316 | 16039 | 10933 |
| | MIMIC-development | 188 | 1744 | 1177 |
| | MIMIC-test | 373 | 3331 | 2243 |
| | UW-test | 518 | 4903 | 3249 |
| UF Health notes | Cancer cohort-training | 503 | 10506 | 3765 |
| | Cancer cohort-test | 126 | 2687 | 1027 |
| | Opioid cohort-test | 100 | 2217 | 754 |

**Task formulation**

We described a clinical context sequence as $X = \{x_1, x_2, \ldots, x_n\}$ and a triplet as $\{\pi = (e_1, r, e_2) | e_1 \in E_1, e_2 \in E_2, r \in R\}$, where $E_1$ and $E_2$ are pre-defined clinical concept types, $R$ is the predefined relation categories, $n$ denotes the sequence length and $x_i$ is the i-th word in the sequence. When considering the applications in cross-domain settings, given the model trained using the source dataset, the goal of transfer learning for clinical concept and relation extraction is to extract all triplets $\pi$ from $X$ in the target dataset. Note that the triplets can share the same clinical concept or relations, i.e., the nested or overlapping concepts. For the model training on the source dataset, we integrated the soft prompt-based learning strategy into two types of LLM architectures. Specifically, encoder-only models formulate the tasks as a two-stage MRC task, where each category of clinical concept and relation is characterized by different soft prompts, and all concepts and relations are extracted from the original input text by prompting the language model; decoder-only models formulate the tasks as a text generation problem, where the concepts and relations are generated directly by the model based on the prompts. Similarly, for the transfer learning on the target dataset, all concepts and relations are inferenced by the soft prompts learned in the training.

**LLMs with P-tuning for SDoH extraction**

This study explored both encoder-only LLMs based on the BERT architecture and decoder-only LLMs based on the GPT architecture through P-tuning. Inspired by P-tuning v2,[35] we implement P-tuning using deep prompt turning, where the trainable prompts were added to all transformer layers in addition to the input layer. For both encoder-only and decoder-only LLMs, we froze the parameters of pretrained LLMs (keeping LLMs parameters unchanged during the training process, i.e., frozen LLMs) and only updated the parameters of soft prompts.

For the encoder-only LLMs, we implemented the soft prompt-based MRC framework for clinical concept extraction and relation extraction. In the first stage, the MRC model is trained to identify the trigger concepts (e.g., "Employment" in the 2022 n2c2 dataset, and "Ethnicity" in the cancer cohort dataset). Specifically, a series of prompt tokens $P = \{p_1, p_2, \ldots, p_m\}$ ($m$ is the length of soft prompt) for each category of trigger concept is initialized to be continuous vector and prepended to the right of the input text $X = \{x_1, x_2, \ldots, x_n\}$ ($n$ is the length of given text) and each transformer layer, then the soft prompt parameter can be updated through model training and the trigger concepts can be identified by the model. In the second stage, the model is trained in a similar manner to identify the attribute concepts and the relations based on the extracted trigger concepts. We introduced a verbalizer to mark the trigger concept spans using two anchor tokens (i.e., [S] and [E]). For example, if the extracted "Drug" concept is "Pantoprazole", we add two anchor tokens [S] and [E] around the concept, resulting in "[S] Pantoprazole [E]" within the input sequence. In these two stages, the final hidden representations of each context token are used for the entity span prediction, we adopt the strategies developed in our previous study,[33] including two binary classifiers to predict the start and end indexes, and a separate classifier to match the start index to the corresponding end index.

For the decoder-only LLMs, we approached clinical concept extraction and relation extraction using a unified text-to-text learning through P-tuning. The soft prompts were initialized as continuous parameters and prepended to the input text and through all the model layers. During P-tuning, only the soft prompt parameters are updated while the LLM parameters were unchanged. We reformulate the extraction of SDoH as a text-to-text learning so that both SDoH concepts and relations can be generated using a unified generative LLM. We convert the gold-standard annotations of training samples into natural language sequences using predefined templates. For the task of clinical concept extraction, we convert the gold label <entity_type_1, entities,..., entity_type_n, entities,> (*n* is the number of entity category) to the format "the extracted [entity_type_1] entities are [entities]; the extracted [entity_type_2] entities are [entities];…, the extracted [entity_type_n] entities are [entities]". For example, "the extracted drug entities are Iodine, Prochlorperazine; the extracted ADE entities are nausea, vomiting." For the task of clinical relation extraction, we convert the golden label <entity_1, entity_2, relation> to the format "the relation between [entity_1] and [entity_2] is [relation]". If there are multiple relations for an input text, we sort them according to their order of appearance and use semicolons to concatenate them.

**Cross-institution and cross-disease transfer learning**

We examined the transfer learning ability of LLMs for two different settings including (1) cross-institution, where we trained models using the MIMIC-III training set and tested the performance using the UW test set, (2) cross-disease domains, where we trained LLMs using clinical notes of cancer patients and tested the performance using clinical notes from patients with opioid use.

**Experimental settings**

We explored two encoder-only LLMs including BERT and GatorTron and a decoder-only generative LLM, GatorTronGPT.

BERT: a bidirectional transformer-based encoder language model pre-trained over a large general English domain corpus. BERT adopted the masked language modeling (MLM) and next-sentence prediction (NSP) training objectives to create deep representations capturing contextual information. BERT is the first transformer model proposed in NLP, which has been widely used for many clinical NLP tasks.

GatorTron[36]: a clinical LLM based on the BERT architecture developed in our previous work, which was pre-trained from scratch using 90 billion words of text (including >82 billion words of de-identified clinical text). We examined GatorTron models with different sizes including 345 million, 3.9 billion, and 8.9 billion parameters.

GatorTronGPT[37]: a generative clinical LLM based on based on the GPT architecture developed in our previous work, which was trained using 277 billion words of text comprising 82 billion words of clinical text and 195 billion words of diverse general English text. We explored GatorTronGPT models with different sizes including 5 billion and 20 billion parameters.

We developed prompt-based models using the open library released in our previous study.[33] We adopted a grid search strategy to optimize hyperparameters, including the learning rate, the training batch size, and the training loss weight. We fine-tuned the transformer models using the training datasets. The best models were selected according to the cross-validation performances measured by micro-averaged strict F1-score. All experiments were conducted using 8 Nvidia A100-80G GPUs.

**Results**

**Table 2** compares the cross-institution performance of GatorTron models developed using P-tuning (i.e., GatorTron-345M-Ptuning, GatorTron-3.9B-Ptuning and GatorTron-8.9B-Ptuning), GatorTronGPT models developed using P-tuning (i.e., GatorTronGPT-5B-Ptuning and GatorTronGPT-20B-Ptuning), GatorTron-345M model developed using traditional fine-tuning without prompts, and BERT-large model developed using traditional fine-tuning without prompts. All models are trained using the MIMIC-III training set and tested using the UW test set. For clinical concept extraction, all P-tuning models except the GatorTron-345M-Ptuning outperformed GatorTron-345M and BERT-large trained using traditional fine-tuning. Among the five P-tuning models, GatorTronGPT-20B-Ptuning achieved the best F1 scores of 0.8378, which outperformed traditional BERT and GatorTron-345M by 9.7% and 8.9%, respectively. GatorTron-345M-Ptuning obtained the lowest F1-score of 0.7379. The performance improved remarkably when increasing the model size from 345 million parameters to over 3.9 billion parameters. For end-to-end extraction of SDoH concepts and relations, all five P-tuning models achieved remarkable performance improvement in F1 scores compared with fine-tuning models. GatorTronGPT-20B-Ptuning achieved the best F1 score of 0.7703, outperforming traditional BERT and GatorTron-345M by 20.4% and 21.8%, respectively. For both tasks, we observed consistent performance improvements by scaling up the size of LLMs.

**Table 2.** Comparison of GatorTron/GatorTronGPT with P-tuning to GatorTron and BERT using fine-tuning for cross-institution applications.

| Model | Training set | Test set | Clinical concept extraction | | | End-to-end Clinical relation extraction | | |
|---|---|---|---|---|---|---|---|---|
| | | | Precision | Recall | F1 | Precision | Recall | F1 |
| BERT-large | n2c2-MIMIC-train | n2c2-UW-test | 0.7559 | 0.7208 | 0.7413 | 0.5821 | 0.5489 | 0.5612 |
| GatorTron-345M | | | 0.7590 | 0.7261 | 0.7488 | 0.5792 | 0.5405 | 0.5526 |
| GatorTron-345M-Ptuning | | | 0.7425 | 0.7276 | 0.7379 | 0.6097 | 0.6236 | 0.6149 |
| GatorTron-3.9B-Ptuning | | | 0.8336 | 0.8109 | 0.8297 | 0.7299 | 0.7098 | 0.7266 |
| GatorTron-8.9B-Ptuning | | | 0.8345 | 0.8114 | 0.8299 | 0.7335 | 0.7102 | 0.7280 |
| GatorTronGPT-5B-Ptuning | | | 0.8219 | 0.8423 | 0.8320 | 0.7598 | 0.7620 | 0.7611 |
| GatorTronGPT-20B-Ptuning | | | 0.8267 | 0.8519 | **0.8378** | 0.7655 | 0.7721 | **0.7703** |

**Table 3.** Comparison of GatorTron/GatorTronGPT with P-tuning to GatorTron and BERT using fine-tuning for cross-disease applications.

| Model | Training set | Test set | Clinical concept extraction | | | End-to-end Clinical relation extraction | | |
|---|---|---|---|---|---|---|---|---|
| | | | Precision | Recall | F1 | Precision | Recall | F1 |
| BERT-large | UF-Cancer | UF-Opioid | 0.8233 | 0.8111 | 0.8172 | 0.7513 | 0.6513 | 0.6977 |
| GatorTron-345M | | | 0.8289 | 0.8124 | 0.8198 | 0.7521 | 0.6529 | 0.6989 |
| GatorTron-345M-Ptuning | | | 0.8209 | 0.8027 | 0.8141 | 0.8006 | 0.7924 | 0.7961 |
| GatorTron-3.9B-Ptuning | | | 0.8696 | 0.8559 | 0.8597 | 0.8379 | 0.8199 | 0.8298 |
| GatorTron-8.9B-Ptuning | | | 0.8699 | 0.8580 | 0.8605 | 0.8386 | 0.8210 | 0.8305 |
| GatorTronGPT-5B-Ptuning | | | 0.8710 | 0.8693 | 0.8708 | 0.8310 | 0.8423 | 0.8389 |
| GatorTronGPT-20B-Ptuning | | | 0.8715 | 0.8760 | **0.8749** | 0.8354 | 0.8519 | **0.8436** |

**Table 3** compares P-tuning models with fine-tuning models for SDoH extraction in a cross-disease application, where the models are trained using UF Health clinical notes of cancer patients and tested using clinical notes from patients with opioid use. For SDoH concept extraction, P-tuning models with over billions of parameters outperformed fine-tuning models in terms of F1 score, except the GatorTron-345M-Ptuning. Among the five P-tuning models, GatorTronGPT-20B-Ptuning achieved the best F1 scores of 0.8749, which outperformed BERT and GatoTron-345M by 5.8% and 5.5%, respectively. The two decoder-only models GatorTronGPT-5B-Ptuning and GatorTronGPT-20B-Ptuning achieved better F1 scores with improvements of 1.0 ~ 6.1% compared with the three encoder-only models. For end-to-end SDoH concept and relation extraction, all P-tuning models outperformed fine-tuning models. GatorTronGPT-20B-Ptuning achieved the best F1 score of 0.8436, outperforming BERT and GatorTron by 14.6% and 14.5%, respectively. The two decoder-only models, GatorTronGPT-5B-Ptuning and GatorTronGPT-20B-Ptuning, consistently outperformed the three encoder-only models by 0.1% ~ 4.8% in F1 score.

**Discussion**

The progress in LLMs has greatly improved patient information extraction including clinical concept extraction and relation extraction. However, most existing methods based on the fine-tuning strategy have limited transfer learning ability to support cross-institution and cross-disease domain applications, as discovered in our previous studies. This study explored P-tuning using two types of transformer architectures, including encoder-only LLM - GatorTron, and decoder-only LLM – GatorTronGPT, for SDoH extraction from clinical narratives in cross-institution and cross-disease domain scenarios. We systematically evaluated P-tuning/fine-tuning with encoder-only/decoder-only LLMs using a cross-institution benchmark dataset and a cross-disease dataset developed using UF Health clinical notes. The experimental results show that the GatorTronGPT-20B-Ptuning model achieved the best F1 score. This study demonstrates the better transfer learning ability of generative clinical LLMs with P-tuning for SDoH extraction for cross-institution and cross-disease applications.

In our previous study,[38] we discovered remarkable performance drops (9.6~10.4% for SDoH concept extraction, 17.5~18.4% for end-to-end SDoH extraction) when fine-tuned BERT and GatorTron using the MIMIC-training set and tested the performance using the UW-test set, indicating that the fine-tuning LLMs have limited transfer learning ability for cross-instruction applications. In another study[34], we also observed remarkable performance drops (8.8% for SDoH concept extraction) when fine-tuned BERT and GatorTron using UF Health clinical notes from cancer patients and tested the performance using the clinical notes from patients with opioid uses. These two studies suggest that fine-tuning LLMs has limited transfer learning ability for cross-institution and cross-disease applications due to the variations in documenting SDoH in EHRs. In our previous study, we examined a traditional solution to annotate additional data and continue fine-tuning using the dataset in the targeted domain and alleviated this issue. In this study, we systematically examined P-tuning with both encoder-only and decoder-only LLMs and demonstrated that P-tuning of generative LLMs has better transfer learning ability for cross-institution and cross-disease applications. The experimental results show that our P-tuning GatorTronGPT models remarkably reduced the performance drops for SDoH concept extraction (3.1% drop for cross-institution settings, 2.2% drop for cross-disease settings) and end-to-end SDoH extraction (4.0% drop for cross-institution settings, 4.5% drop for cross-disease settings), outperforming traditional models. Traditional fine-tuning LLMs follow a sequence labeling solution for concept extraction and formulate relation extraction as a classification task over candidate concept pairs, which has limited transfer learning ability tied to specific datasets and tasks. Whereas P-tuning of generative LLMs solves SDoH extraction using a unified text-to-text learning architecture and uses soft prompts to control information extraction, which has better transfer learning ability for cross-institution and cross-disease applications.

P-tuning has more improvements for the end-to-end SDoH extraction (13.0% ~ 14.4% increase in F1 score) compared with SDoH concept extraction (6.5 ~ 8.2% increase in F1 score). One potential reason is that the errors that happen in the concept extraction are amplified in the relation classification where the concept pairs for classification are typically generated by enumerating the combinations among detected concepts. Generative LLMs with Ptuning solved the end-to-end extraction of SDoH using a unified text-to-text learning architecture, which achieves better performance for concept extraction and prevents error amplification. Decoder-only LLMs GatorTronGPT achieved better performance compared with the encoder-only LLMs GatorTron, indicating that generative LLMs benefit more from P-tuning. We observe that GatorTron-345M-Ptuning has a significant performance gap compared with fine-tuned GatorTron-345M and other P-tuning models, indicating that P-tuning requires large-size LLMs as the parameters of LLMs are frozen and only the soft prompts can be updated. Our results show that P-tuning of larger LLMs with billions of parameters has better transfer learning abilities to support cross-domain NLP applications.

This study has limitations. We focus on information extraction of SDoH from clinical narratives, future studies should examine more clinical NLP tasks in cross-domain scenarios. This study tests transfer learning from a source dataset to a target dataset using the same task, future studies should examine complex scenarios where the source task and target task are different. We will explore complex cross-domain scenarios in clinical NLP and develop computationally efficient P-tuning solutions.

**Conclusion**

This study demonstrates the transfer learning ability of generative LLMs for cross-institution and cross-disease clinical NLP applications through P-tuning. However, employing larger LLMs with over billions of parameters is necessary to benefit from P-tuning.

**Acknowledgments**

We would like to thank the n2c2 challenge organizers for providing the annotated corpus. We acknowledge the support from the Cancer Informatics Shared Resource in the UF Health Cancer Center. We gratefully acknowledge the support of NVIDIA Corporation and the NIVIDA AI Technology Center (NVAITC) UF program. We would like to thank the UF Research Computing team, for providing computing power through UF HiPerGator-AI cluster.

**Funding statement**

This study was partially supported by a Patient-Centered Outcomes Research Institute® (PCORI®) Award (ME-2018C3-14754), a grant from the National Cancer Institute, R01CA246418, grants from the National Institute on Aging, NIA R56AG069880, R01AG080624, R01AG083039, R01AG080991, R01AG084236, R01AG084178,

R01AG076234, National Institute of Allergy and Infectious Diseases, NIAID R01AI172875, National Heart, Lung, and Blood Institute, R01HL169277, National Institute on Drug Abuse, NIDA R01DA050676, the Cancer Informatics Shared Resource supported by the UF Health Cancer Center and the UF Clinical and Translational Science Institute Biomedical Informatics Program. The content is solely the responsibility of the authors and does not necessarily represent the official views of the funding institutions.